\documentclass[11pt, oneside]{article}   	% use "amsart" instead of "article" for AMSLaTeX format
\usepackage{geometry}                		% See geometry.pdf to learn the layout options. There are lots.
\usepackage{graphicx}				% Use pdf, png, jpg, or eps§ with pdflatex; use eps in DVI mode

\newcommand{\R}{\mathbb{R}}

\newcommand{\E}{\mathbb{E}}
\usepackage{amssymb,amsmath,amsfonts,latexsym}
\usepackage{bbm}
\usepackage{mathrsfs}
\usepackage{color}

\title{People Mover's Distance: Class level geometry using fast pairwise data adaptive transportation costs}
\author{Alexander Cloninger\thanks{Applied Mathematics Program, Yale University}  \and 
	Brita Roy\thanks{Section of General Internal Medicine, Department of Medicine, Yale School of Medicine} \and 
	Carley Riley\thanks{Division of Critical Care Medicine, Department of Pediatrics, Cincinnati Children’s Hospital Medical Center} \and 
	Harlan M. Krumholz\thanks{Center for Outcomes Research and Evaluation, Yale University}}
\date{}							% Activate to display a given date or no date

\begin{document}
\maketitle

\begin{abstract}
	We address the problem of defining a network graph on a large collection of classes.  Each class is comprised of a collection of data points, sampled in a non i.i.d. way, from some unknown underlying distribution.  The application we consider in this paper is a large scale high dimensional survey of people living in the US, and the question of how similar or different are the various counties in which these people live.  We use a co-clustering diffusion metric to learn the underlying distribution of people, and build an approximate earth mover's distance algorithm using this data adaptive transportation cost. 
\end{abstract}

\section{Introduction}
We consider the problem of building accurate distances between two classes of points $Z_i$, where each class is defined only by its members $Z_i = \{x_j \in \R^m: class(x_j) = i\}$.  Each member $x_j$ can be thought of as a person described by $m$ observed features (e.g. questions in a survey, electronic health record data).  Building such class distances requires the ability to accurately measure similarity and dissimilarity between the two point clouds $Z_i$ and $Z_j$.  The purpose of these comparisons is to build a weighted graph of similarity between all classes $\{Z_i\}_{i=1}^C$, and use several eigenfunctions of this graph adjacency matrix to build a low-dimensional set of vectors that characterize the similarity and dissimilarity of all $C$ classes simultaneously \cite{diffusion2006}.

The problem of comparing distributions between point clouds becomes very difficult when the dimension of the data points $\{x_j\}\subset \R^m$ is large.  Classical statistical methods based on mean and covariance matrices of the responses in a class are rarely able to capture more complex descriptors such as multimodal distributions or outliers due to masking effect, especially in high dimensional data \cite{highDimStat}.  Also,  summary statistics break down for distribution matching in situations where the population is non-normally distributed or high-dimensional \cite{highDimDist,duda2012pattern}.

One method that is capable of measuring distances between two high-dimensional distributions is \emph{earth movers distance} (EMD) \cite{imageEMD}.
%use a more complex distance between distributions known as \emph{earth movers distance} (EMD), which is capable of measuring distances between two general, high-dimensional distributions \cite{imageEMD}.  
Informally, earth movers distance can be thought of the minimal amount of work necessary to turn all points $\{x_k\}\subset Z_i$ into points $\{y_\ell\}\subset Z_j$ under a cost of movement $d(x_k,y_\ell)$.  As an example of people in a survey, earth movers distance would be similar to  the sum of all distances from a person $x_i\in Z_i$ to the closest person $y_\ell \in Z_j$, under the constraint that two $x_i$ cannot map to the same $y_\ell \in Z_j$.  

Earth mover's distance has been used in various forms in image analysis \cite{imageEMD}, audio signal classification \cite{audioEMD}, and general database measurements \cite{databaseEMD}.  In this paper, we use these methods for a new application; organizing counties by patterns of residents' survey responses that provide more information than summary statistics and factor analysis.

We consider an application of this approach to organizing US counties according to responses of individuals in those counties taking the 2014 Gallup-Healthways Well-Being Index (GHWBI) survey \cite{Healthways}.  The GHWBI is a national telephone survey of adults from every state in the US, collecting responses from around $200,000$ people a year. The survey was designed to measure a variety of dimensions of overall well-being, such as physical and mental health, healthy behaviors, pride in the community, supportive relationships, sense of meaning and purpose in life, and how life is going overall.  There are $56$ questions in the 2014 survey.  These questions represent five general domains, 
and the overall Well-being Index score is calculated as an average of these five domains.  These domains are Purpose (liking what you do and motivation to achieve your goals), Social (supportive relationships and love in your life), Financial (managing your economic life to reduce stress and increase security), Community (safety, happiness, and pride in your community), and Physical (good health behaviors and outcomes) \cite{Healthways}.  Along with this, the covariance matrix of questions has rank much greater than 5, so there exist sub-categories as well.  These questions give rise to many different types of individuals' well-being profiles.  
For this application, we aim to compare and contrast the counties' health and happiness by the distribution of residents' well-being profiles living in that county.  

\begin{figure}[!h]
	\begin{center}
		\begin{tabular}{cc}
			\includegraphics[width=.4\textwidth]{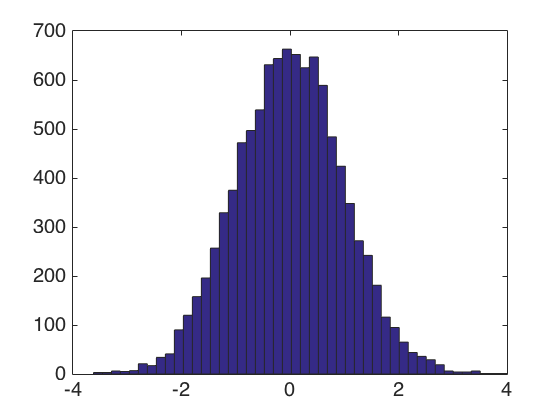} & 
			\includegraphics[width=.4\textwidth]{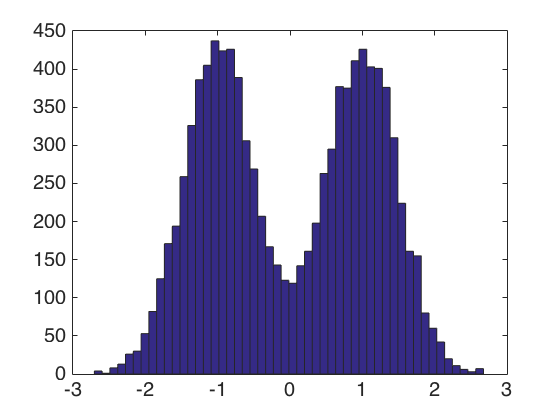}
		\end{tabular}
		\includegraphics[width=.9\textwidth,height=.2\textwidth]{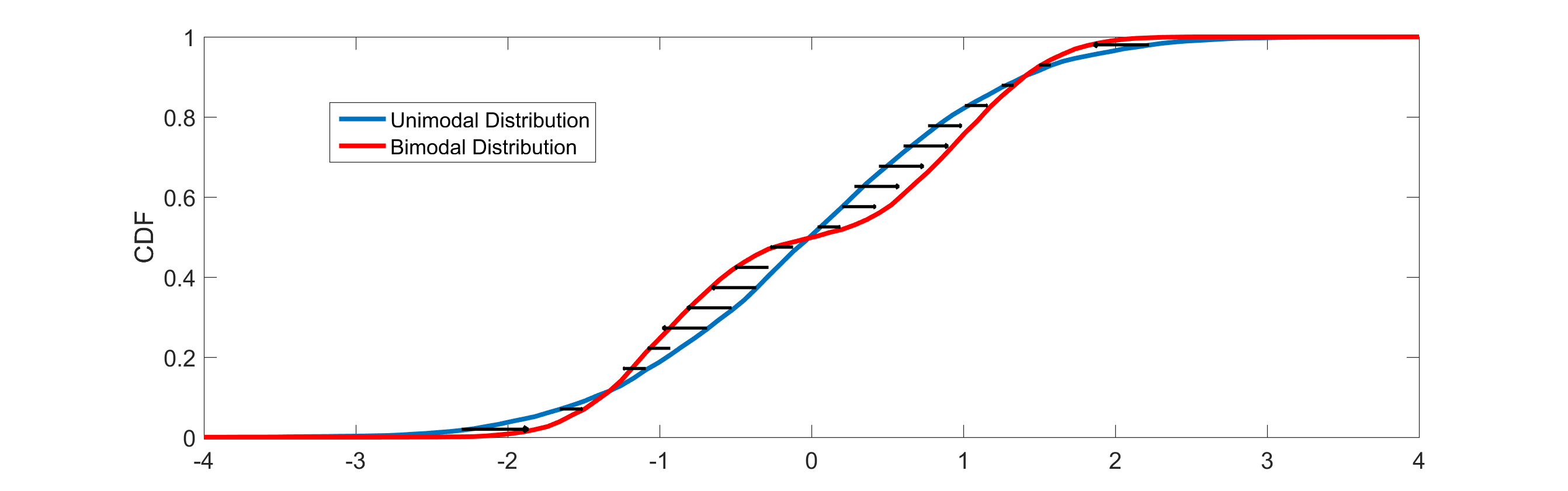}
	\end{center}
	\caption{Two hypothesized distributions of WBI score with same mean and variance, (top left) everyone in county is mostly average, (top right) there are clearly two types of people in this county with disparate WBI levels, (bottom) CDFs of two distributions with black arrows denoting the which people must be ``shifted'' from the unimodal distribution to match the bimodal distribution.}\label{fig:twoDists}
\end{figure}

As an example of the benefits of earth movers distance, consider the distributions in Figure \ref{fig:twoDists} showing the possible distribution of two counties' Well Being Index score.  While these distributions have the same mean and standard deviation, they are clearly dissimilar and represent significantly different health profiles.  If we instead consider how far we'd have to move individual people along the WBI score (increasing or decreasing their individual WBI score) to transform a unimodal distribution into a bimodal distribution, it's clear that the distance will be quite large.  

%Earth mover's distance yields more accurate measure of similarity between distributions than moment matching.  Characterizing counties by their mean response in a survey can fail to capture the true distance between two counties, especially in situations where the population is non-normally distributed.  

There are two main difficulties to defining this type of distance, one on the point/individual level and one on the class/county level.  First, on the point level, the collection of all points $X = \{x_i\}\subset \R^m$ (e.g. people that responded to the GHWBI survey) does not come with an a priori distance metric $d(x,y)$.   Subsets of features of each person may be highly correlated, or behave differently in different subsets of people.  For this reason, we define a bigeometric organization of the people's profiles to account for such feature interdependencies, and construct a diffusion metric between the people.  We discuss this method in detail in Section \ref{bigeom}.  

The second difficulty comes in computing the class level earth mover's distance (e.g. EMD distance between counties) from $d(x,y)$.  A major hurdle to using EMD is it's $O(N^3 \log N)$ computational complexity for an $N$ bin histogram.  Because it is required to calculate every pairwise distance between classes, computation time becomes a major hurdle.  To surmount the barrier,  we instead focus on an approximate earth mover's distance based on multi scale tree distance, as originally proposed by \cite{leebEMD,JacobsEMD}, which can be computed in linear time.  The use of localized binning to create a distance metric has been used in a variety of contexts, such as in the linguistic bag-of-words models \cite{bagOfWords} and topic models \cite{topicModel}, in which synonyms are grouped together before defining distances between collections of words.  We discuss this method in Section \ref{approxEMD}.

\subsection{Language and Notation}
Throughout the paper we will refer to data points $(x_i, z_i)$ and a class of data points $Z_j = \{x_i : z_i =j\}$, a set constituted by a specific collection of data points.   In the context of the GHWBI survey, class $Z_j$ corresponds to a county $j$ made up of the people in that county $\{x_i : z_i =j\}$.  

For the rest of the paper, we will refer interchangeably to classes of points as \emph{counties} (i.e. Prince George's County, Los Angeles County, etc), and the data points $x_j\in \R^m$ as \emph{people} that responded to the $m=56$ question GHWBI survey and live in a given county.

We will also refer to types of points.  These correspond to small clusters of people that share similar features (i.e. $\|x_i - x_j\|_2$ is small).  Types can be defined at different resolutions, depending on the number of types needed to break up the space.  In the context of GHWBI, these types can be thought of as different types of well-being profiles.

\section{Algorithm}
\subsection{Approximating EMD from a Metric}\label{approxEMD}

Let each data point be described by $(x_i, z_i)$, where $x_i \in X\subset \R^{N\times d}$ describes the features of the datapoint, and $z_i \in \{1, ..., K\}$ is the class in which $x_i$ belongs.  We wish to characterize the $K$ classes $Z_j = \{x_i : z_i =j\}$ by defining an  earth mover's distance between the classes
\begin{equation}\label{eq:EMD}
\begin{split}
EMD(Z_i, Z_j) &= \sup_f \int f(x) \left( p_{Z_i}(x) - p_{Z_j}(x) \right) dx \\
\textnormal{such that } & \left| f(x) - f(y) \right| \le d(x,y)^\alpha,
\end{split}
\end{equation}
where 
$$p_{Z}(x_k) = \frac{1}{|Z|} \begin{cases} 1, & x_k \in Z \\ 0, & otherwise \end{cases}.$$

%In most applications, knowledge of a distance metric $d:X\times X\rightarrow \R^+$ is a priori unknown.  While there are many possible methods for building an unsupervised metric $d$ on this data, we choose to focus on diffusion metrics which measure the low frequency changes in the data.  Moveover, because 

\begin{figure}[!h]
\begin{center}
\includegraphics[width=.6\textwidth]{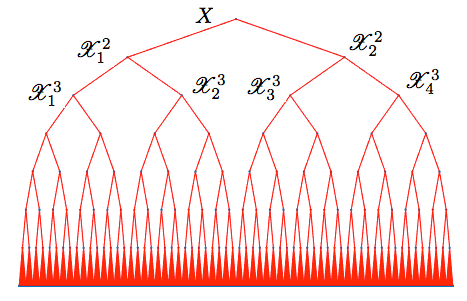}
\end{center}
\caption{Breakdown of $Y$ into folders.}\label{fig:treePlot}
\end{figure}

We defer choice of $d:X\times X \rightarrow \R^+$ to Section \ref{bigeom}, and focus here on fast approximations to EMD.  We focus on the tree approximation to EMD, which is strongly equivalent to the true EMD for any metric $d(x,y)^\alpha$ for $\alpha<1$ \cite{leebEMD}.  Leeb builds this approximation by constructing a partition tree on the feature space, given that it is equipped with a distance metric $d: Y\times Y\rightarrow \R^+$.   A partition tree on $Y$ is a sequence of $L$ tree levels $\mathscr{Y}^\ell$, $1\le \ell\le L$.  Each level $\ell$ consists of $n(\ell)$ disjoint sets $\mathscr{Y}^\ell_i$ such that 
 \begin{eqnarray*}
 Y = \bigcup_{i=1}^{n(\ell)}  \mathscr{Y}^\ell_i.
 \end{eqnarray*}
 Also, we define subfolders (or children) of a set $\mathscr{Y}^\ell_i$ to be the indices $I^{\ell+1}_i \subset\{1, ..., n(\ell+1)\}$ such that
 \begin{eqnarray*}
 \mathscr{Y}^\ell_i = \bigcup_{k \in I^{\ell+1}_i} \mathscr{Y}^{\ell+1}_k.
 \end{eqnarray*}
For notation, $\mathscr{Y}^1 = Y$ and $\mathscr{Y}^L_i = \{y_i\}$.  See Figure \ref{fig:treePlot} for a visual breakdown of $Y$.

Now for a distribution $p_i$, denote $\delta_{\ell,k}^i$ to be $p_i$'s average for the questions in folder $\mathscr{Y}_k^\ell$.
Also, let $\gamma_{\ell,k} = \frac{|\mathscr{X}_k^\ell|}{|X|}$. Then the approximate earth mover distance is defined as 
\begin{eqnarray*}
 EMD_{\alpha,\beta}(p_i , p_j) = \sum_\ell 2^{\alpha \ell} \sum_k \gamma_{\ell,k}^\beta | \delta_{\ell,k}^i -  \delta_{\ell,k}^j |.
\end{eqnarray*}

\subsection{Bigeometric Organization for a Metric on the Points}\label{bigeom}

In this section, we define a people distance metric $d(x,y)$ for $x,y\in X$.  The metric is built following the bigeometric organization algorithm described in \cite{coifman2011, ankenman2014, leebEMD}.   We summarize the algorithm here.  For notation, we refer to the transpose $Y = X^\intercal\in \R^{d\times N}$ to consider the organization of the features.
 \begin{enumerate}
 \item Define an affinity between the features $K(y_i, y_j) = e^{-\|y_i - y_j\|/2\sigma^2}$ and compute a diffusion embedding via the first $m_Y$ eigenvectors, which we denote $\Phi_t(Y)$.  
 
 \item Construct a partition tree on $\Psi_t(Y)$ that we denote $\mathscr{Y} = \{\mathscr{Y}^\ell\}_{\ell=1}^L$.  
 
 To construct the tree on the embedded points $\Psi_t(Y)\subset \R^d$, the bottom folders $\mathscr{Y}^{L-1}$ are determined by choosing a fixed radius $\epsilon$ and covering $\Psi_t(Y)$ with balls of radius $\epsilon$.  Each subsequent level of the tree is then generated as combinations of the children nodes that are ``closest'' together under the distance $\|\Psi_t(y_i) - \Psi_t(y_j)\|_2$.

 \item Construct a kernel matrix $K$ on the points $X$ using an earth mover distance from the partition tree on the points $X$.  For a point $x_i$, we consider it as a distribution over the questions $Y$, and compute the EMD between it and another point $x_j$ via
 \begin{eqnarray*}
 EMD_{\alpha,\beta}(x_i , x_j) = \sum_\ell 2^{\alpha \ell} \sum_k \gamma_{\ell,k}^\beta | \delta_{\ell,k}^i -  \delta_{\ell,k}^j |,
 \end{eqnarray*}
 where $\delta_{\ell,k}^i$ is $x_i$'s average response for the questions in folder $\mathscr{Y}_k^\ell$.
The kernel function is then 
 \begin{eqnarray*}
 k(x_i, x_j) =  e^{-\frac{EMD_{\alpha,\beta}(x_i, x_j)}{\epsilon}}.
 \end{eqnarray*}

\item Compute a diffusion embedding of the points via the first $m_X$ eigenvectors, which we denote $\Phi_t(X)$.  We finally define a metric $d:X\times X\rightarrow \R^+$ via
\begin{eqnarray}
d(x_i,x_j) &=& \|\Phi_t(x_i) - \Phi_t(x_j) \|_2.
\end{eqnarray}

\item Build a partition tree on $\Phi_t(X)$ that we denote $\mathscr{X} = \{\mathscr{X}^\ell\}_{\ell=1}^L$ in the same way as in Step 2.

\end{enumerate}

\subsection{Class EMD}

We define the distance between two classes to be the approximate EMD between the two classes $Z_i = \{x_k : z_k =i\}$ and $Z_j = \{x_k : z_k =j\}$.  We define the class distance as the distance between data points $\{x_i\}$,  
\begin{eqnarray*}
EMD_{class}(Z_i, Z_j) &=& \sum_{\ell=1}^L 2^{\alpha\ell} \sum_{k=1}^{n_\ell} \gamma_{k,\ell}^\beta \left|\delta_k^\ell(Z_i)   - \delta_k^\ell(Z_j) \right|,
\end{eqnarray*}
where $\delta_k^\ell(Z) = \sum\limits_{x\in \mathscr{X}_k^\ell} p_{Z}(x)$, and $\gamma_{k,\ell} = \frac{|\mathscr{X}_k^\ell |}{|X|}$.

\begin{figure}[!h]
 \begin{tabular}{cccc}
 \includegraphics[width=.2\textwidth]{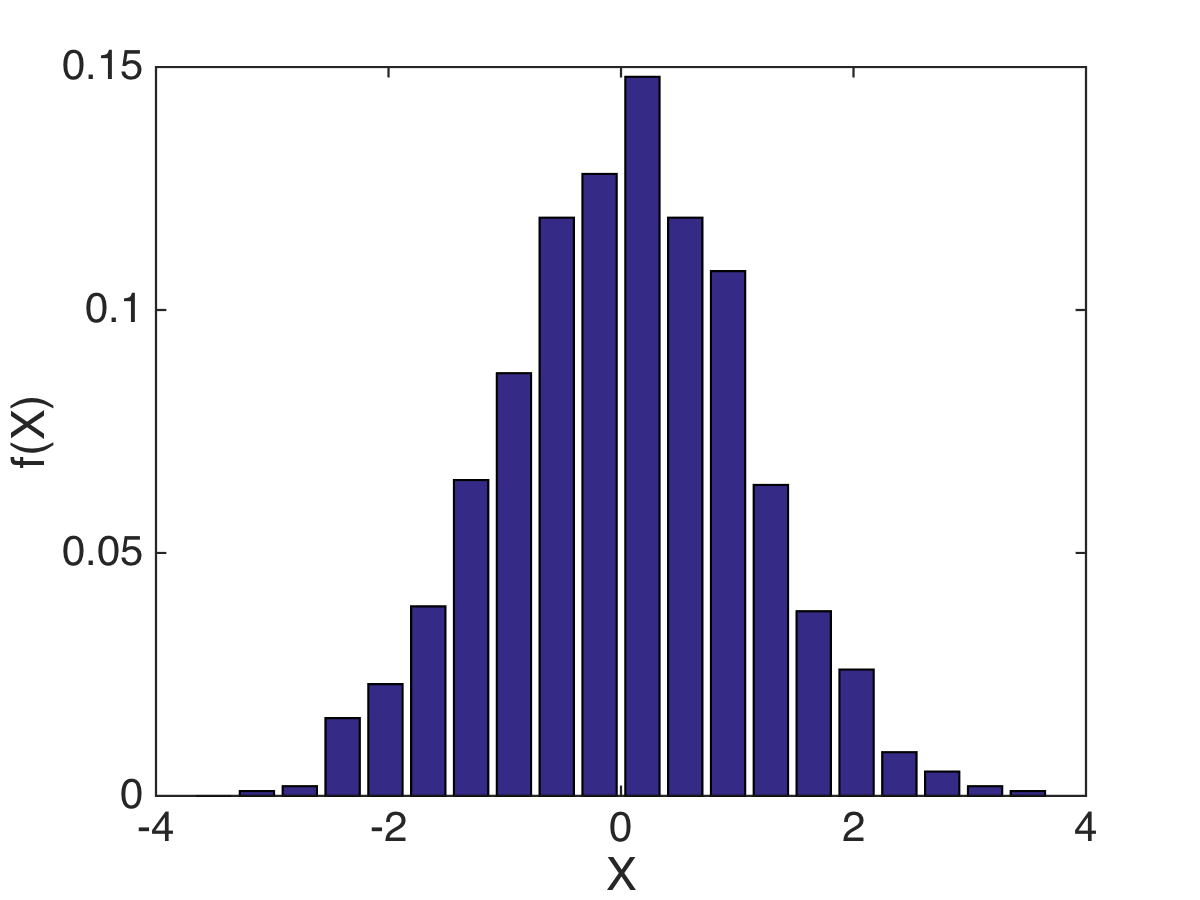} &
      \includegraphics[width=.2\textwidth]{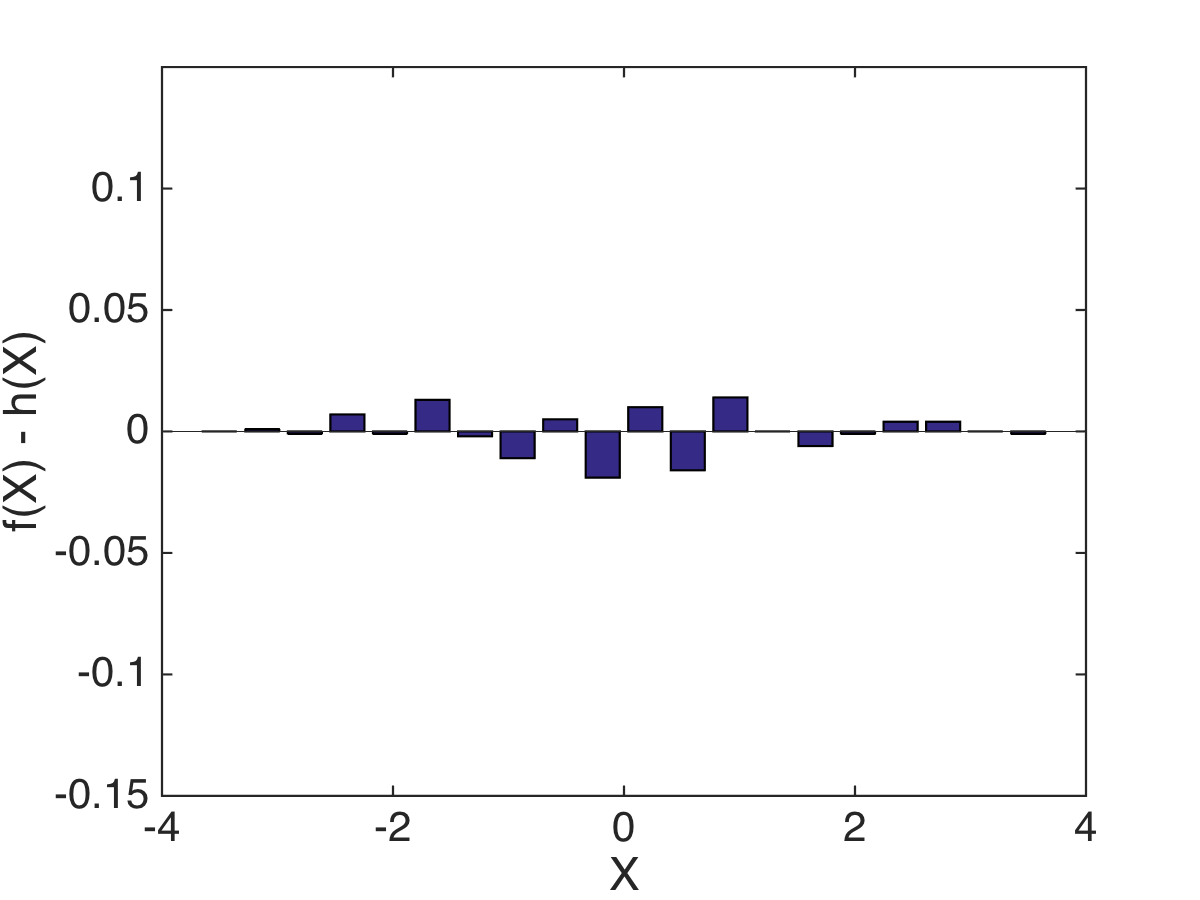} & 
  \includegraphics[width=.2\textwidth]{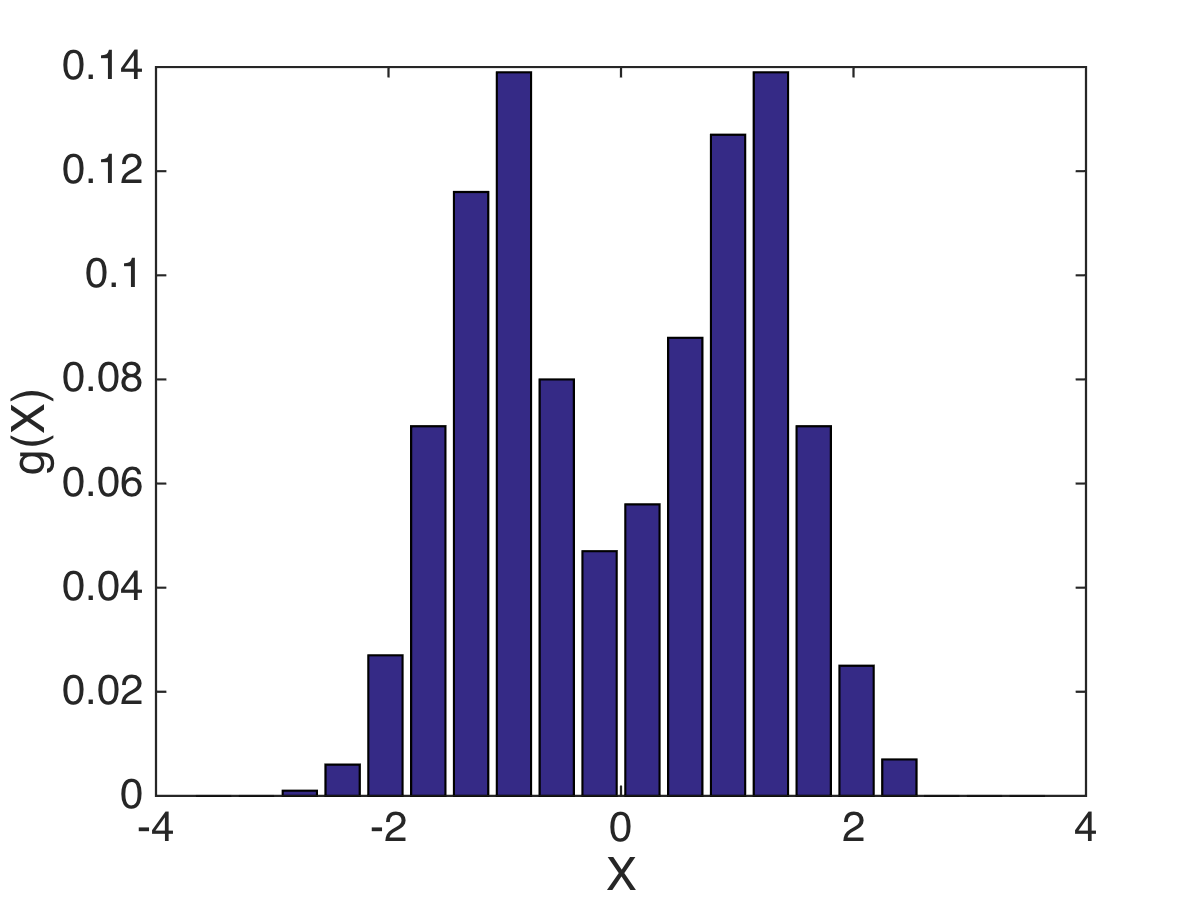} & 
   \includegraphics[width=.2\textwidth]{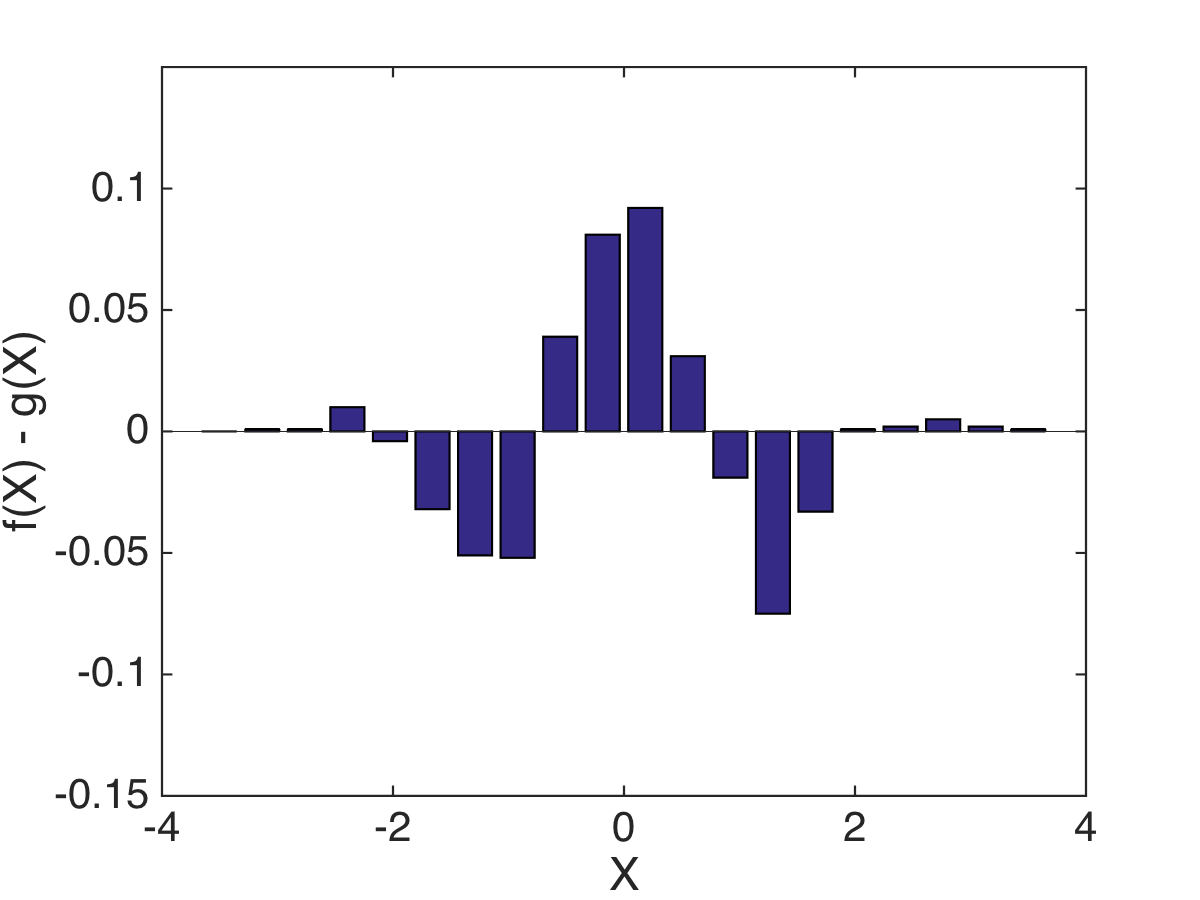} \\
    \includegraphics[width=.2\textwidth]{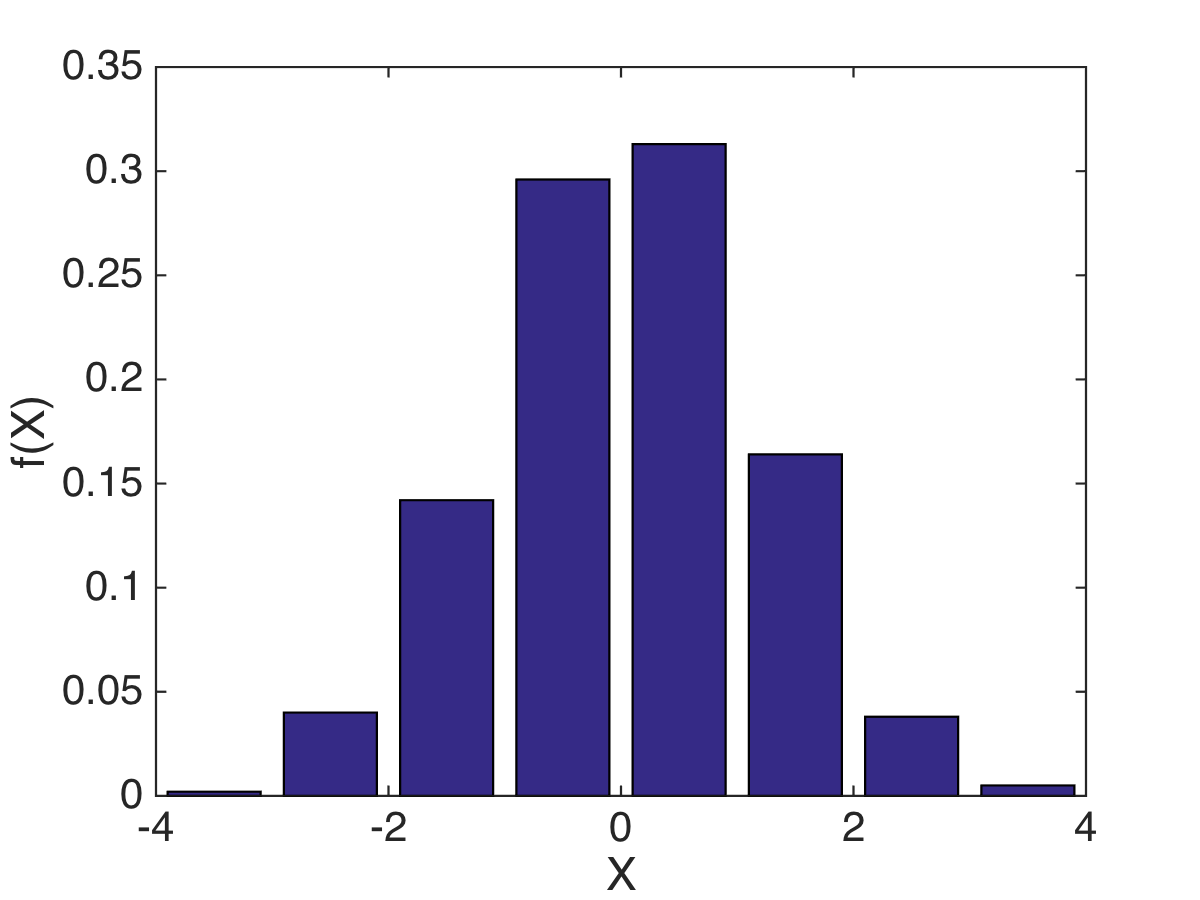} &
        \includegraphics[width=.2\textwidth]{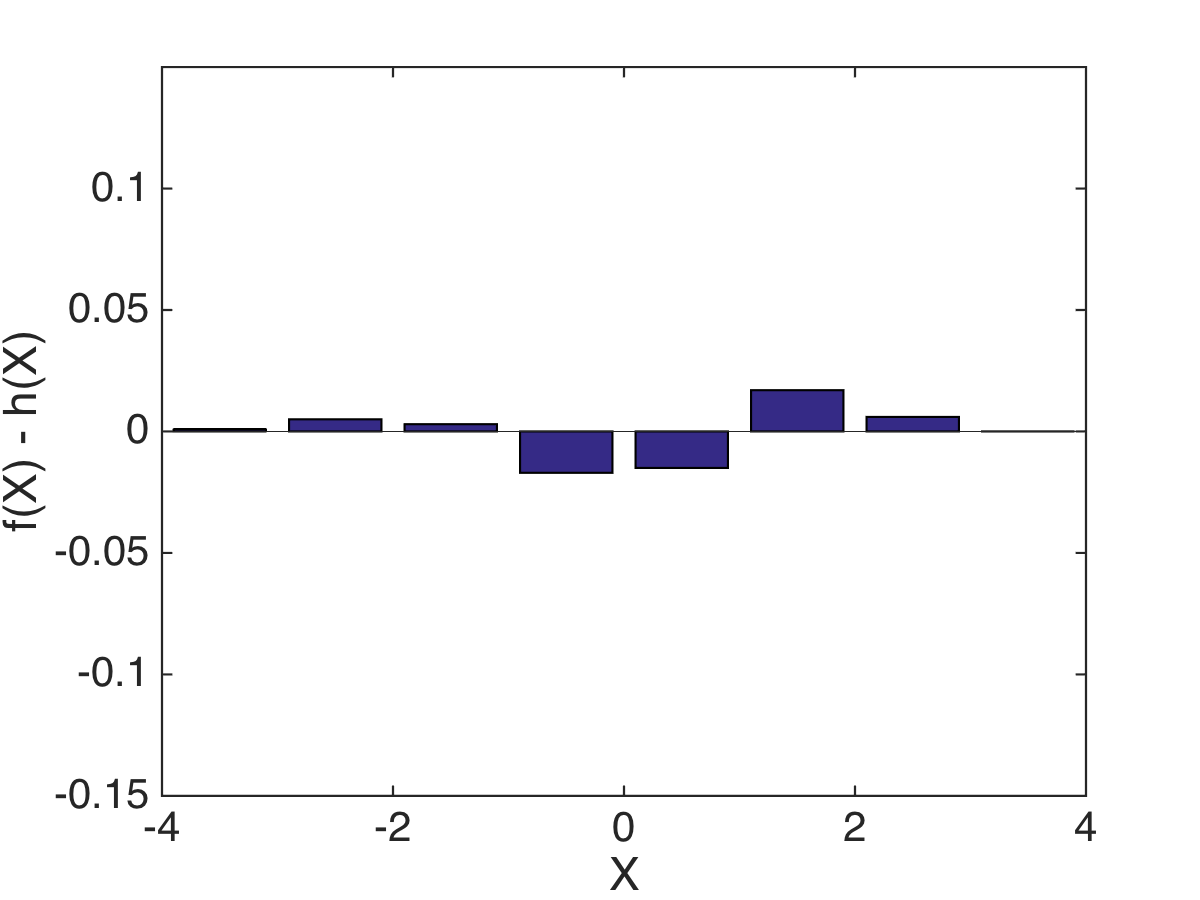} & 
  \includegraphics[width=.2\textwidth]{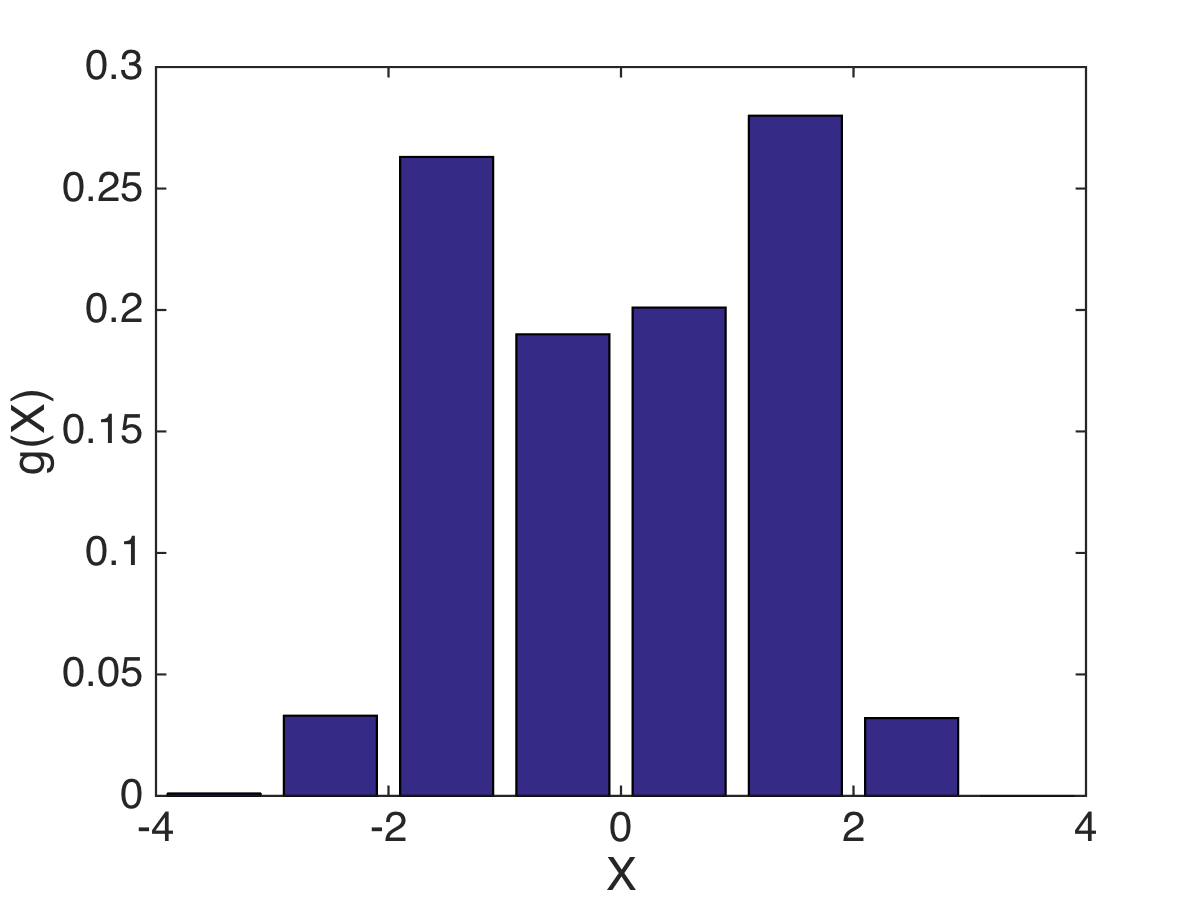} & 
   \includegraphics[width=.2\textwidth]{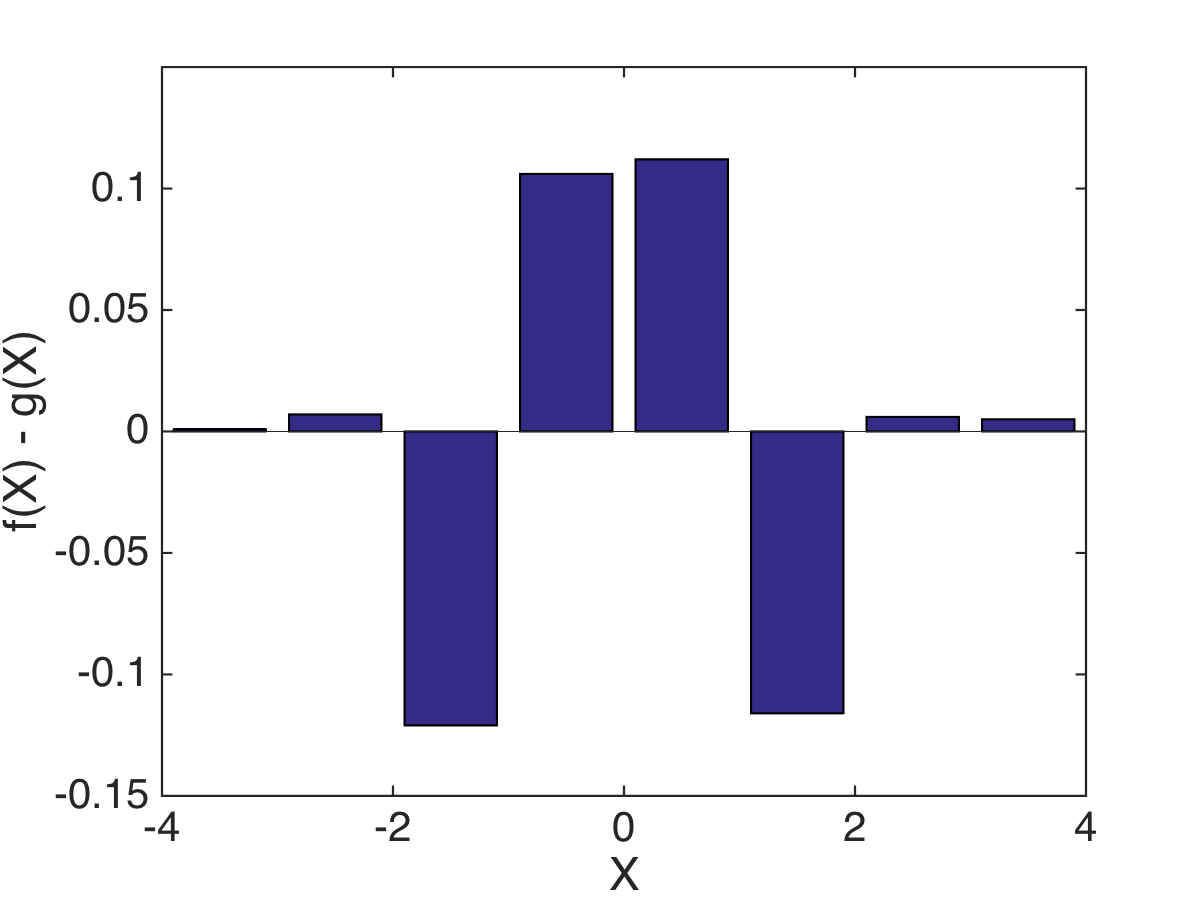} \\
    \includegraphics[width=.2\textwidth]{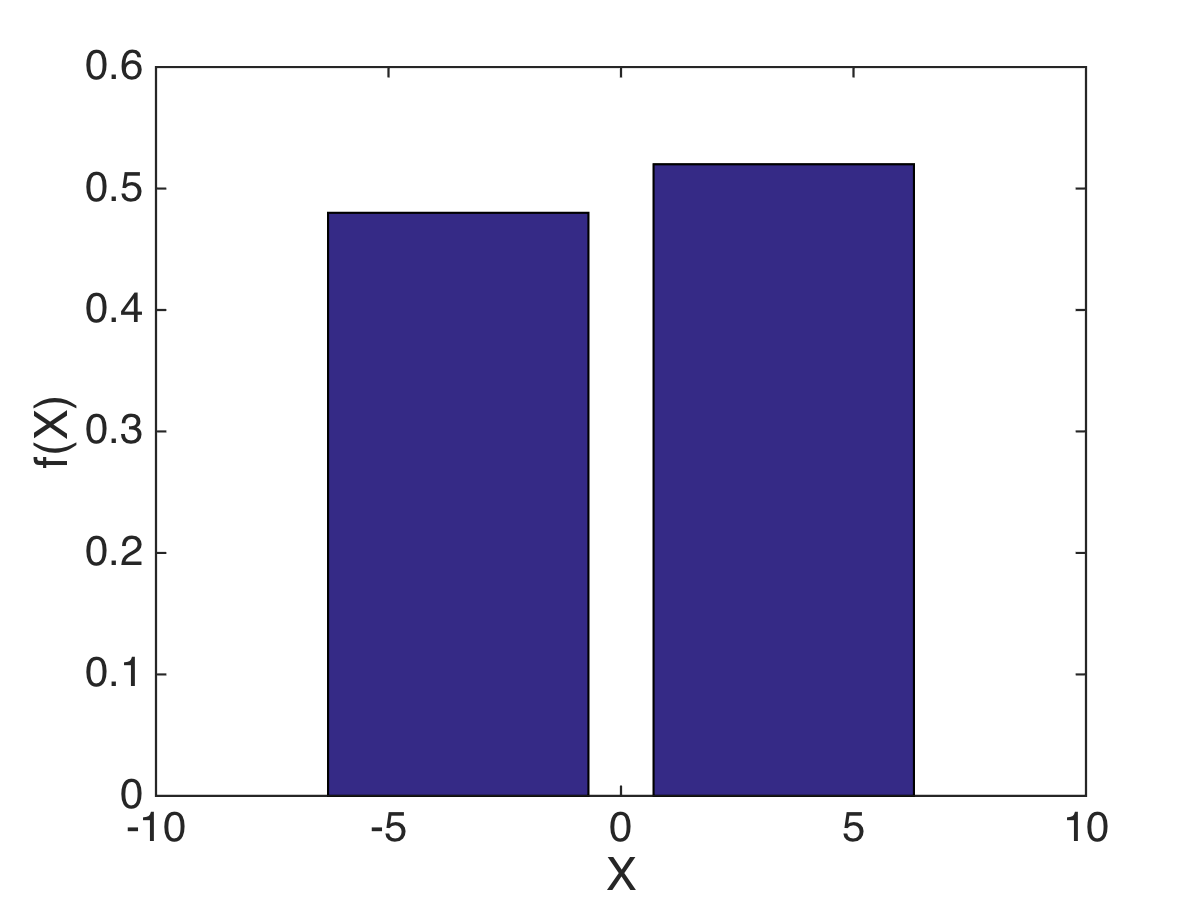} & 
        \includegraphics[width=.2\textwidth]{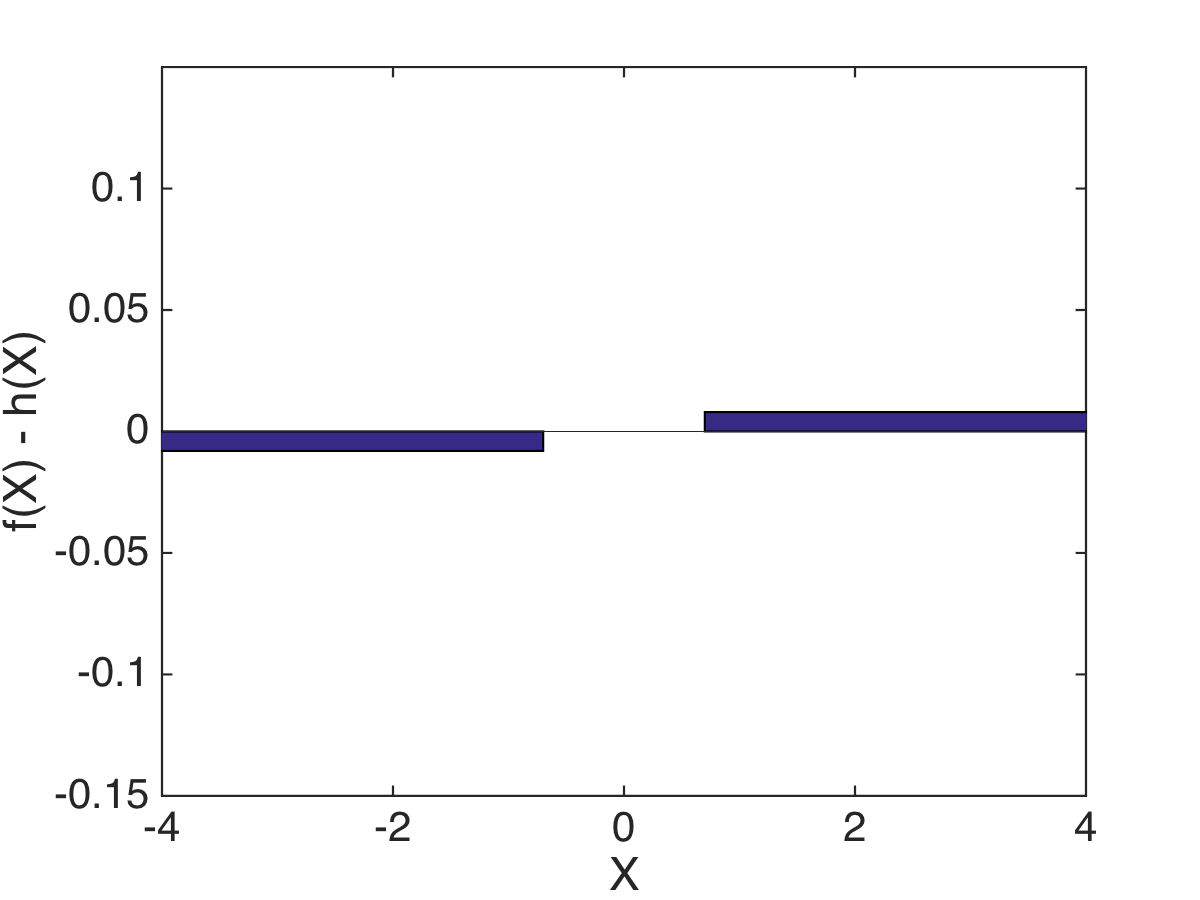} & 
  \includegraphics[width=.2\textwidth]{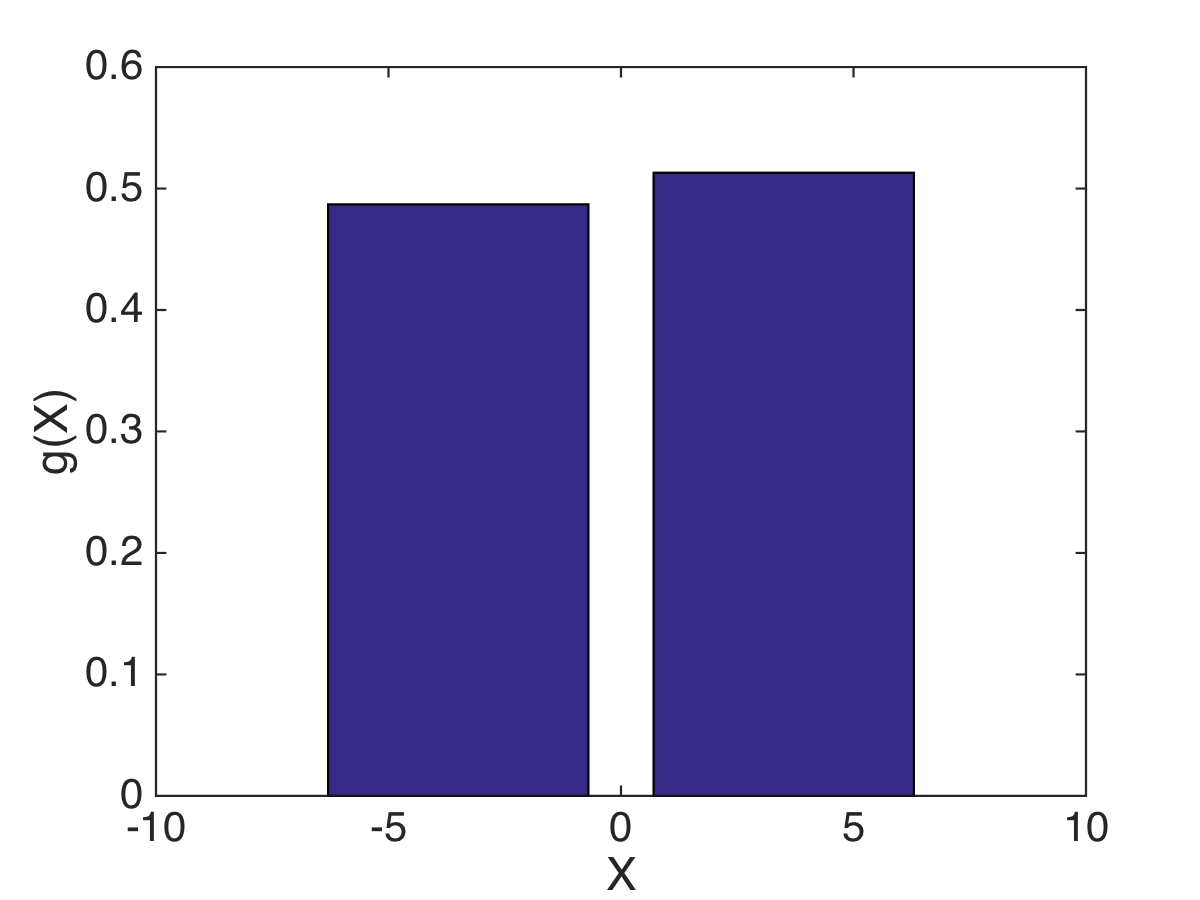} & 
   \includegraphics[width=.2\textwidth]{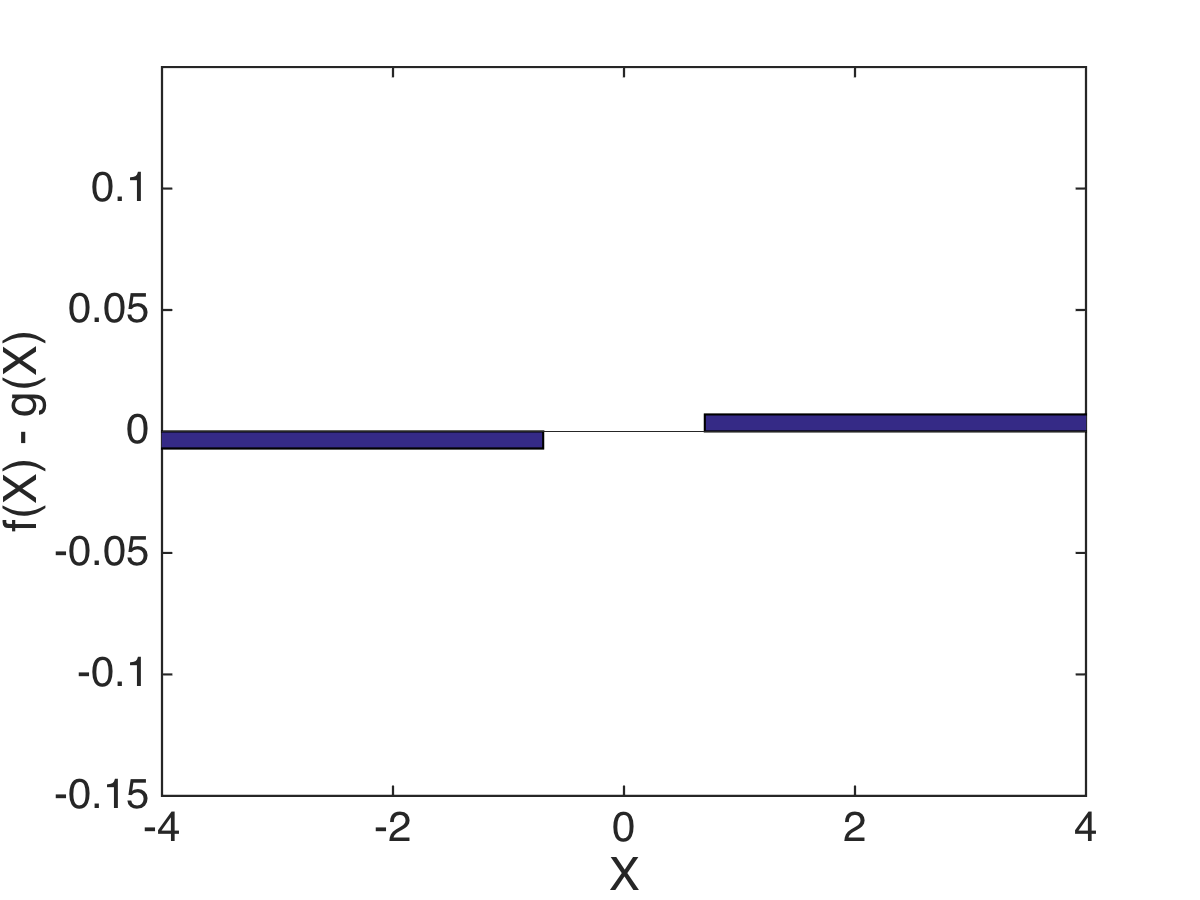} 
   \end{tabular}
   \caption{Histograms of several distributions and differences between histograms.  Top to bottom: 20 bins, 8 bins, 2 bins.  Left to right: one peak normal distribution, difference between two instantiations of one peak normal distribution, two peak distribution, difference between one peak distribution and two peak distribution.}\label{fig:histEMD}
   \end{figure}

This can be interpreted as the amount of work necessary to change one class' members into the other.  
 By building the hierarchical tree of data points, we are clustering the points into ``types'' at various scales of similarity.  At each scale, we characterize a class $Z_i$ by the histogram of ``types'' of points that belong to the class.  See Figure \ref{fig:histEMD} for an example which assumes all people lie on a 1D line of well-being, and two counties have differing profiles at multiple scales.
 
The main benefit of this approach is that it catches the intricacies of the point distribution in a way that the means of the class features don't, and also defines the distance without the need to estimate moments of the features, which would require a large number of points.  

From here, we generate an embedding of the classes which encodes their EMD similarity.  This is done by constructing the graph adjacency matrix between the counties $\{Z_i\}$ via
\begin{eqnarray*}
	K(Z_i, Z_j) = e^{-EMD_{class}(Z_i, Z_j)/sigma},
\end{eqnarray*}
and taking the largest eigenvectors $\Xi_{class} = \begin{bmatrix} \xi_1 & ... & \xi_d  \end{bmatrix}$ of the kernel $K$.  These eigenvectors form a new set of coordinates for the counties that preserves the local neighborhood structure generated by $EMD_class: \{Z_i\} \times \{Z_i\} \rightarrow \R^+$.

\section{GHWBI Organization and Validation}

\begin{figure}[!h]
\begin{center}
\includegraphics[width=.5\textwidth]{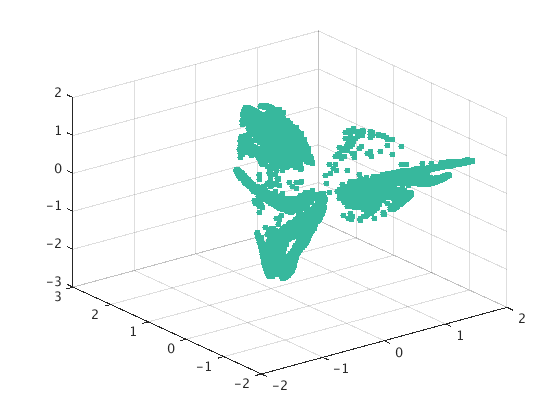} 
\end{center}
\caption{Low frequency eigenvectors for organization of people's survey data.}\label{fig:peopleEmbedding}
\end{figure}

We begin by considering the people distance metric $d(x,y)$ on the Gallup-Healthways Well-being Index survey.  Figure \ref{fig:peopleEmbedding} shows the first three low-frequency eigenfunctions of the well-being adjacency matrix.  This embedding came from 3 iterations of the bigeometric organization algorithm and represents only 3 of the 8 dimensions of $\Phi_t(X)$, and with $\alpha=0.5$ and $\beta=1$.  %\textcolor{red}{We also report the hierarchical organization of the features in Table \ref{tab:peopleFeatures}.}

\begin{figure}[!h]
\begin{center}
\includegraphics[width=.5\textwidth]{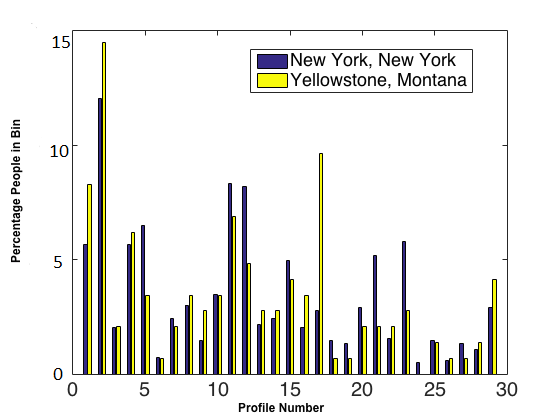} 
\end{center}
\caption{Example histograms for two large counties.}\label{fig:countyHists}
\end{figure}

%Table \ref{tab:peopleClusters} shows one level of the tree of health profiles, where each cluster is represented by the mean centroid of the people in that cluster.  
We use the hierarchical tree of clusters generated by the embedding to construct profiles of the counties at various levels.  The tree breaks down into a number of levels, with 1, 2, 5, 29, and 475 profiles, respectively.  Not all parent nodes have the same number of children as the tree was built bottom up via diffusion distance, as mentioned in Step 5 of Section \ref{bigeom}.  Figure \ref{fig:countyHists} shows examples of these profiles for several counties.  

\begin{figure}[!h]
\begin{center}
\includegraphics[width=.5\textwidth]{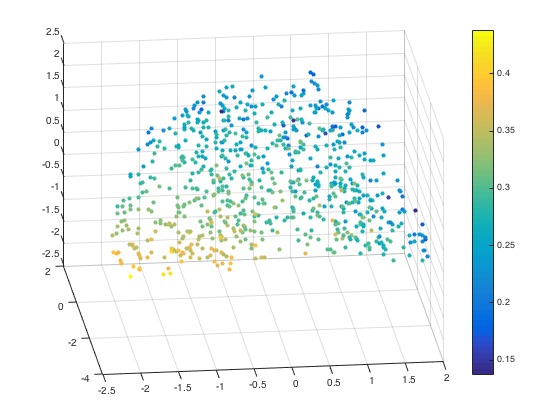} 
\end{center}
\caption{Embedding of 749 largest counties, colored by fraction of county's people in profile $\mathscr{X}_1^5$.}\label{fig:countyEmbedding}
\end{figure}

We only include counties with $\ge50$ respondents, which lowers the number of counties to 749 from the original 3136.   We use levels 2, 3, and 4 of cluster tree, where the levels have 2, 5, and 29 profiles respectively.  We chose the levels due to the sample size of each county, since level 3 of the health cluster tree had 475 bins and the average county only had 186 people sampled in 2014.   These levels are combined with $\alpha=-0.5$, so that more refined clusters are given slightly higher weight.   The embedding of all these counties is shown in Figure \ref{fig:countyEmbedding}, colored by the percentage of people from each county in $\mathscr{X}_1^5$ as an example of a health profile.  

\begin{figure}[!h]
	\footnotesize
	\begin{center}
		\begin{tabular}{ccc}
			\includegraphics[width=.25\textwidth]{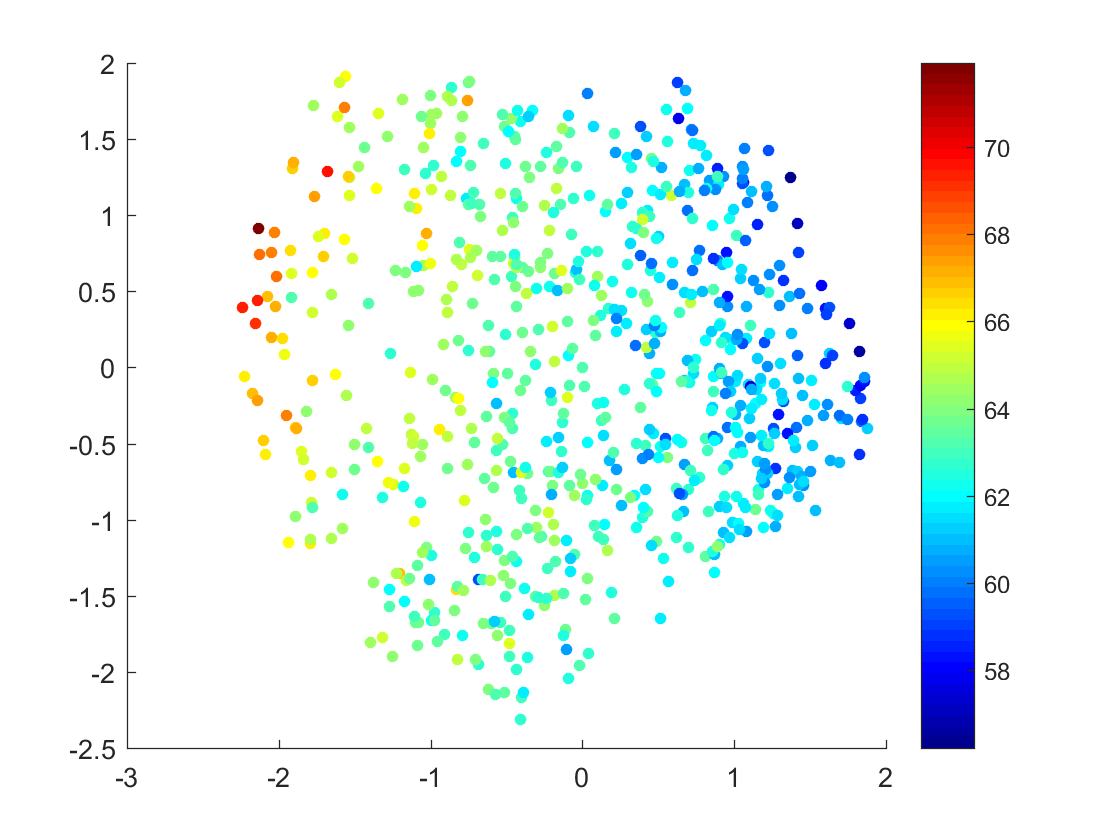} & 
			\includegraphics[width=.25\textwidth]{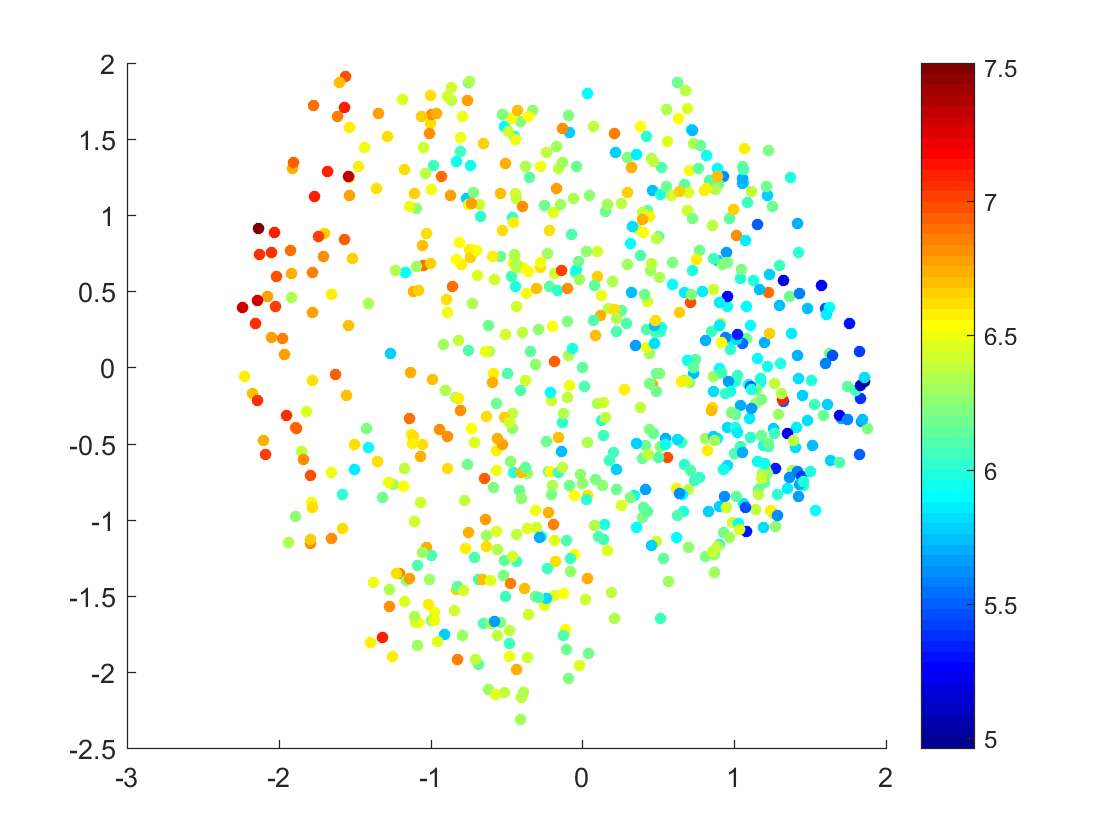} & 
			\includegraphics[width=.25\textwidth]{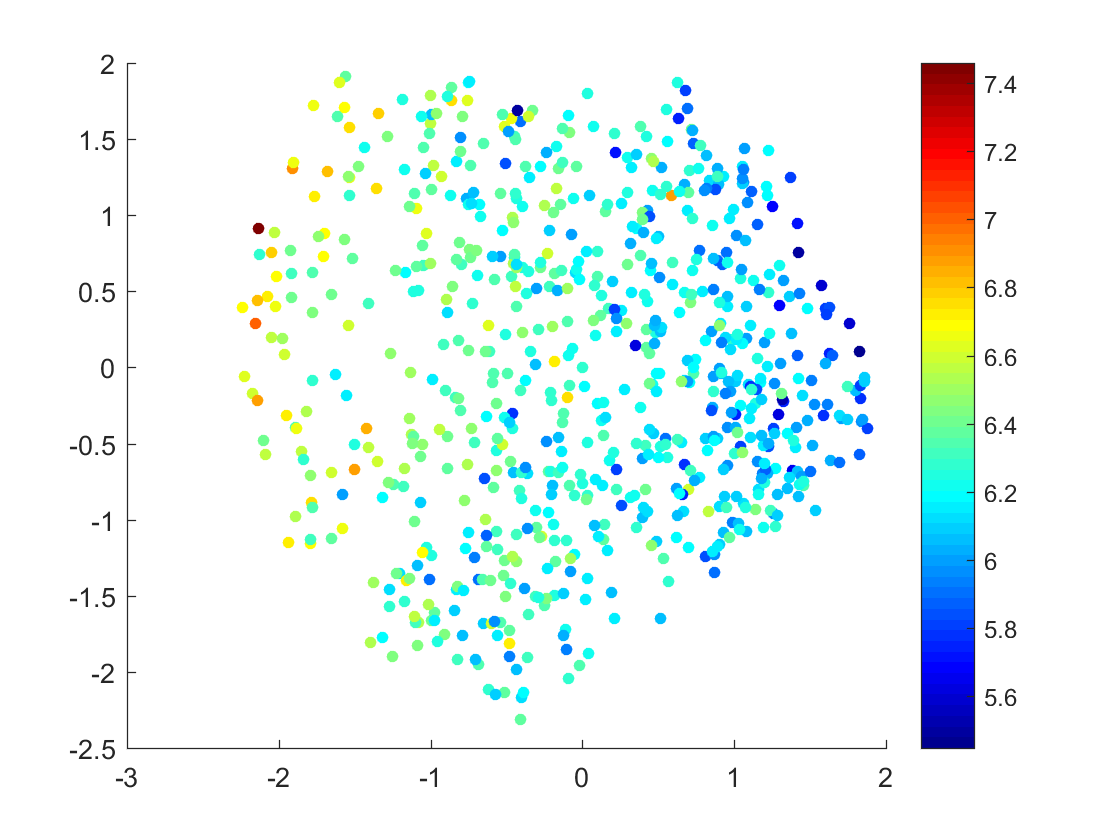} \\
			WBI Score & Community Score & Social Score \\
			\includegraphics[width=.25\textwidth]{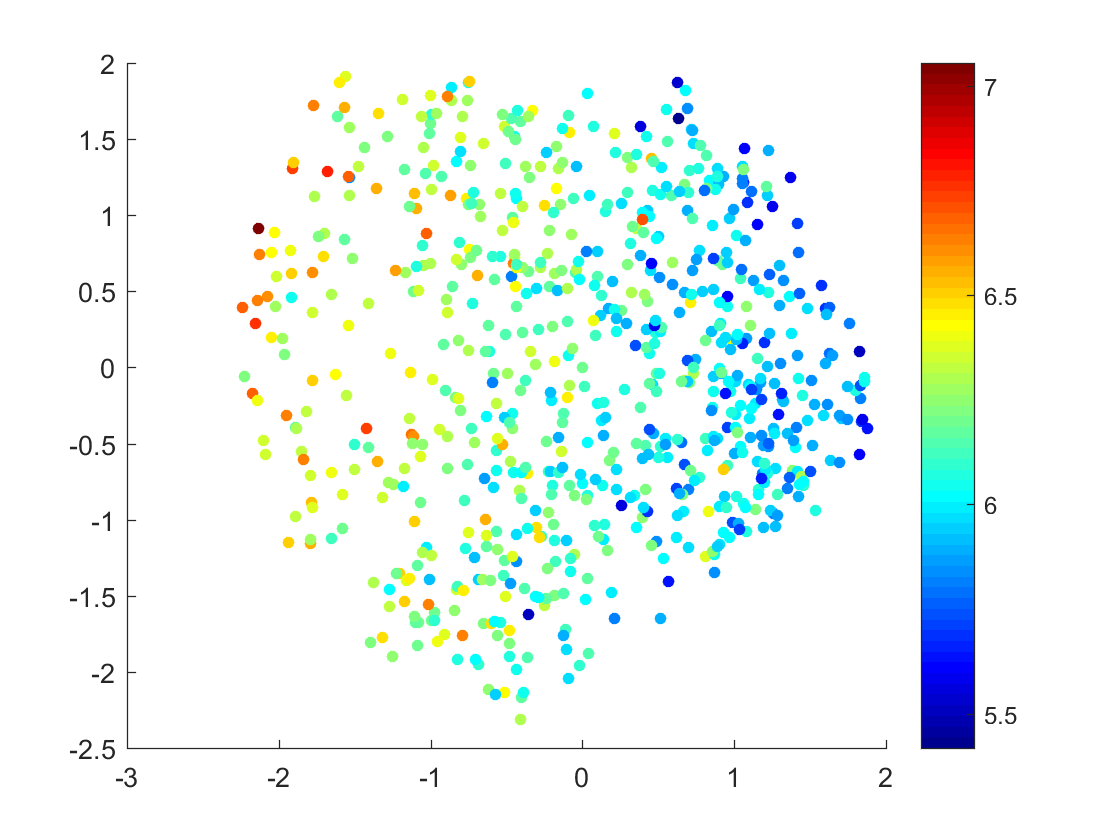} & 
			\includegraphics[width=.25\textwidth]{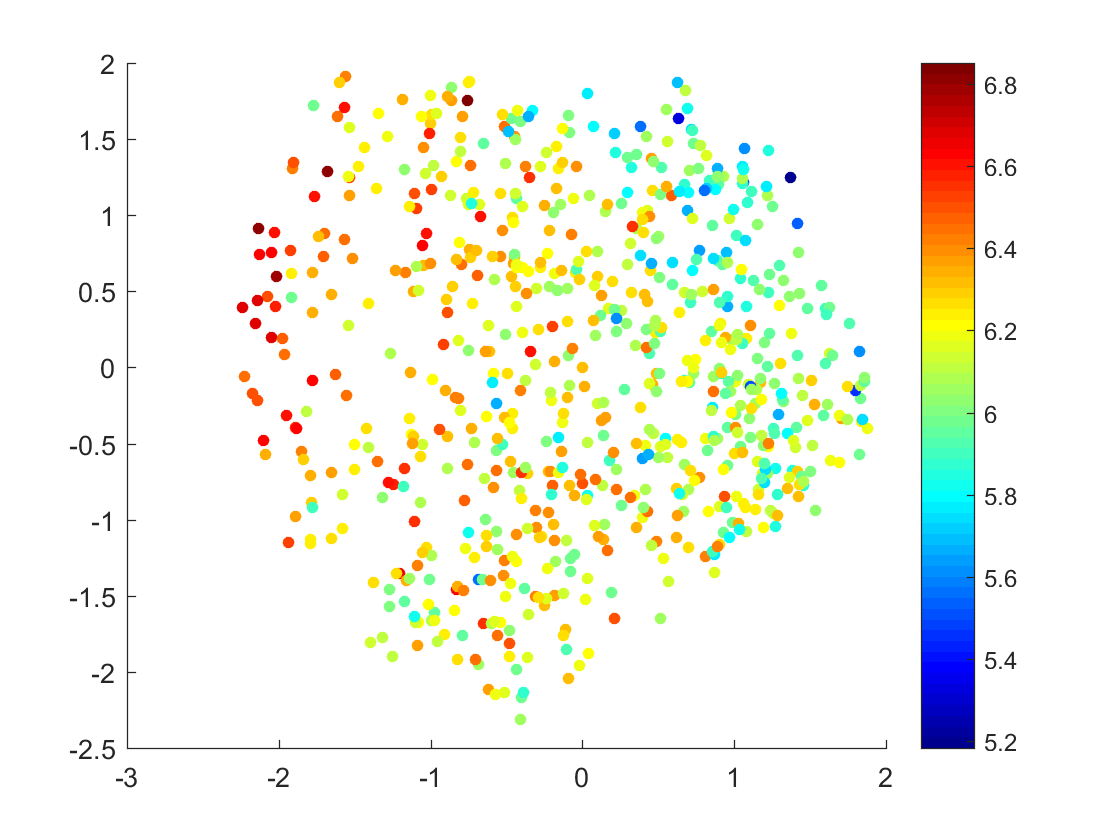} & 
			\includegraphics[width=.25\textwidth]{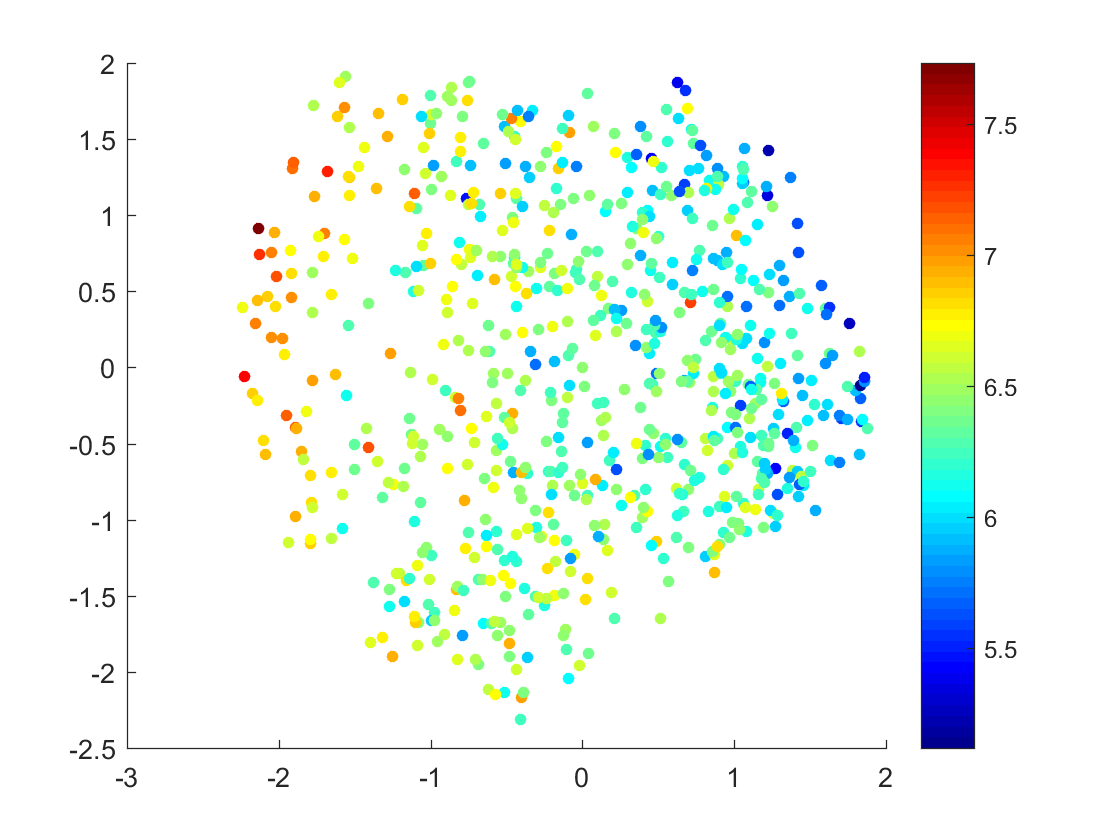} \\
			Purpose Score & Physical Score & Financial Score 
		\end{tabular}
	\end{center}
	\caption{Organization of 749 largest counties, colored by the mean factor value of everyone that lives in that county.}\label{fig:factorColors}
\end{figure}

We also show the embedding of the counties into two dimensions via the eigenvectors $\Xi_{class}$.  The counties colored by the 5 domains of the GHWBI analysis, as well as the overall well being score, in Figure \ref{fig:factorColors}.  For each county $Z_i$, we compute the average score $\E_{x\in Z_i} [x_j]$ for factor $j$.  This shows that our EMD organization of the counties preserves the general trends of the 5 factor domains.  This is not surprising, as both the factors and the EMD organization built in a data adaptive fashion from the original survey responses.  However, this organization gives additional insight, such as easily identifying counties that are outliers with respect to one of the factor domains relative to their overall placement in the EMD organization.  

\begin{figure}[!h]
	\footnotesize
	\begin{center}
	\begin{tabular}{cc}
		\includegraphics[width=.4\textwidth]{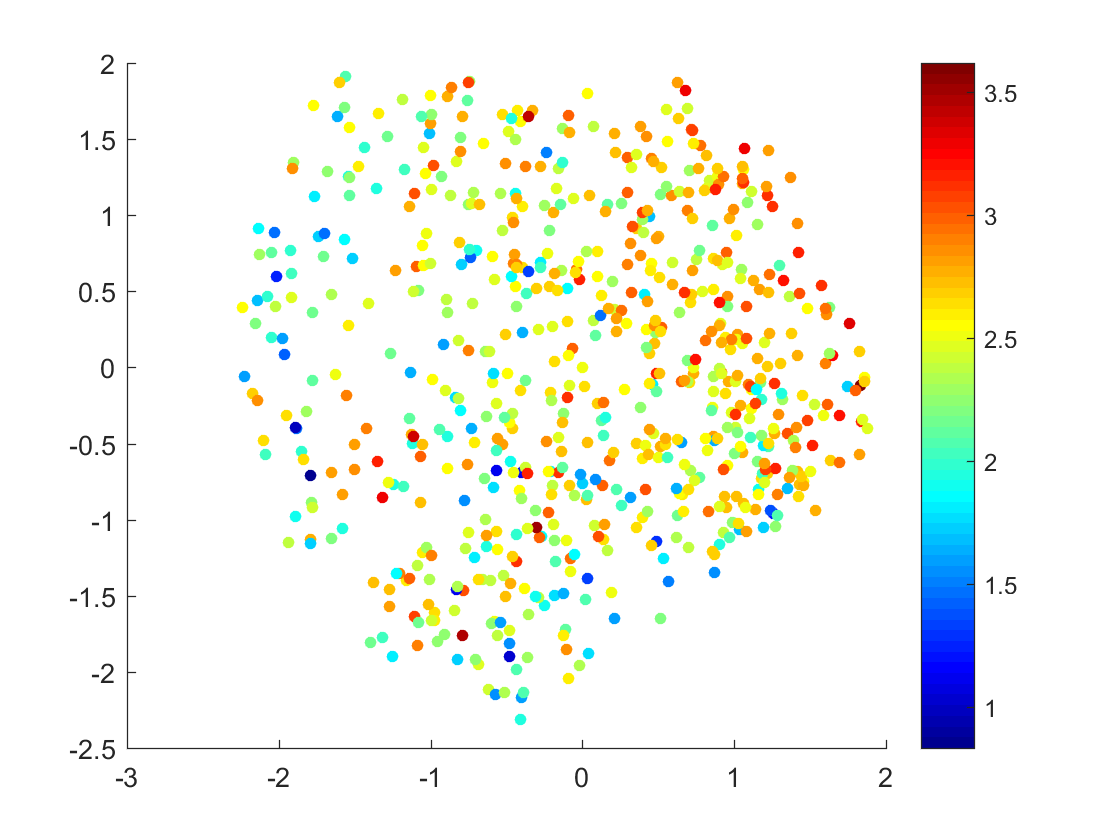} & 
		\includegraphics[width=.4\textwidth]{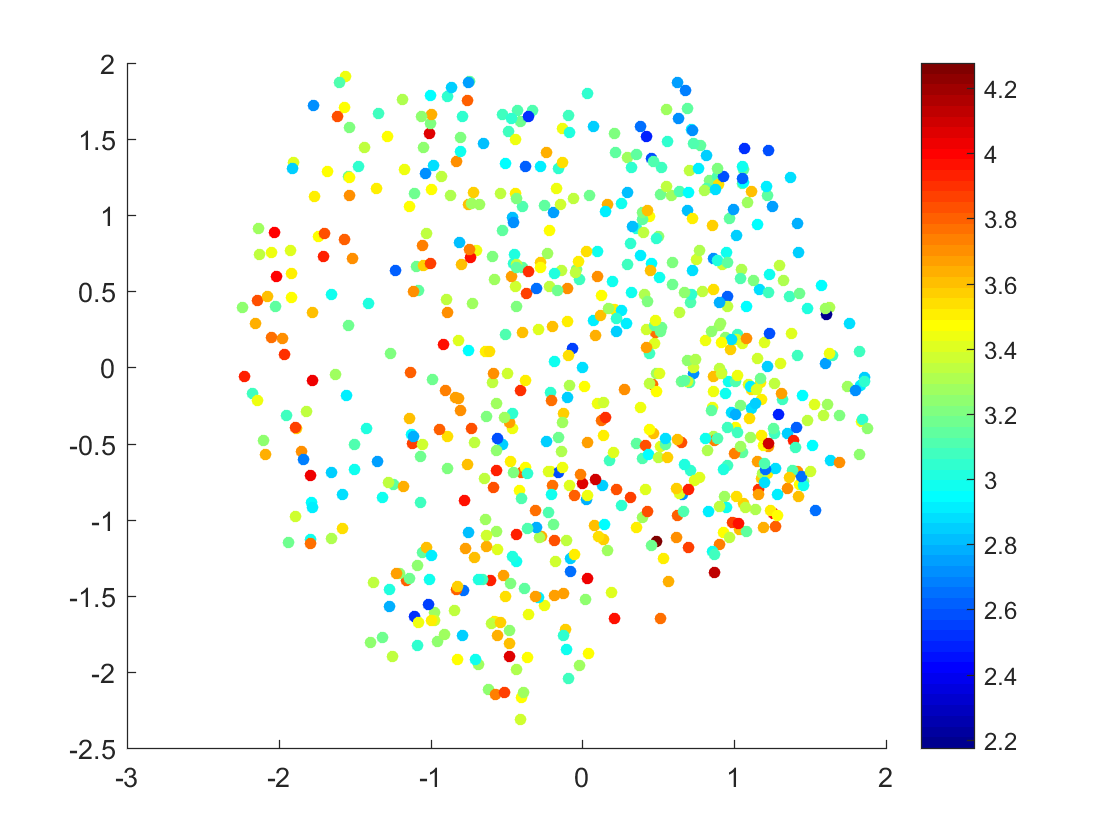} \\
		Log Percent in SNAP & Log Percent with Bachelors degree or higher
	\end{tabular}
	\end{center}
\caption{2D map of large counties colored by 2015 census data.  Colors represent percent of population on log scale for visual ease. }\label{fig:validation}
\end{figure}

Because this embedding is unsupervised, we also need to validate the geometry of the counties by examining whether the geometry has any correlation with external county level 2015 US Census data.  As an example in Figure \ref{fig:validation}, we consider our embedding of large counties that have reported census data, colored by percentage of the county using SNAP (food buying assistance) and colored by the percentage of the county that has attained a Bachelors degree or higher.  These are just two chosen proxies for economic and education status, any county-level statistic could have been chosen as an indicator.

%Unfortunately, we only had access to census data that matched 163 of the 749 largest counties, so the figures are a little more sparse than the complete EMD organization in Figures \ref{fig:countyEmbedding} and \ref{fig:factorColors}.  For this reason, we have reproduced Figure \ref{fig:factorColors} with only those 162 counties showing, and this is displayed in Figure \ref{fig:factorColors_small}.

We see in Figure \ref{fig:validation} that our EMD organization roughly predicts the percentages of people on SNAP or with a strong education, as well as shows the separation level sets of the census percentage.  However, it is not an exact match, which allows for the identification of outliers.  For example, the counties that have a high percentage of people on SNAP are located on the far right of our EMD map, which is the region we've previously identified as counties with both a low overall WBI and a low score for all of the factors.

%In addition, we can see a county $Z_i$ that has a similar well-being profile to counties with high educational achievement, though the rate of educational achievement in $Z_i$ differs significantly.  These types of counties can be easily identified and become the subject of further study.  This means we can now identify counties that have populations with similar experiences of well-being, but varying demographic characteristics.  This outlier information could also be applied to county-level outcomes and other types of county level information.

We also provide a measure for determining whether the clear visual correlations between the county statistics and the embedding in Figure \ref{fig:validation} are significant.  To do this, we construct a row normalized kernel $K_{embedding}$ on the two dimensional county embedding $\Xi_{class}$, which encodes the local neighborhood structure in the embedding.  This is done via
\begin{equation*}
K_{embedding}(Z_i, Z_j) = \frac{exp\left(-\|\Xi_{class}(Z_i) - \Xi_{class}(Z_j)\|_2^2/\sigma^2\right)}{\sum_{k} exp\left(-\|\Xi_{class}(Z_i) - \Xi_{class}(Z_k)\|_2^2/\sigma^2\right) }.
\end{equation*}
Now if some function $f:\{Z_i\} \rightarrow \R$ is generated in a way that depends on the graph structure (i.e. is highly correlated with the organization), then $K_{embedding} f \approx f$.  Or similarly, 
\begin{equation*}
error(f) = \frac{\|f - K_{embedding}f\|}{\|f\|}
\end{equation*}
will be small. 

We use a permutation test to determine whether either function 
\begin{eqnarray*}
f_{SNAP}(Z_i) &=& \textnormal{Percent of residents of $Z_i$ using SNAP},\\
f_{Bachelors}(Z_i) &=& \textnormal{Percent of residents of $Z_i$ with Bachelors degree}, 
\end{eqnarray*}
are significantly correlated with the county organization.  
This is done by permuting the counties with some permutation $\pi$, and comparing the value of $error(\widetilde{f})$ to $error(f)$, where $\widetilde{f}_i = f(Z_{\pi_i})$.  %Note that we chose a log scale for these scores to be more robust to large outlier values (e.g. $60\%$ of people in a county on SNAP when the median is $12\%$).  The following comments also apply to the percentage itself without the log, but are omitted for 

\begin{figure}[!h]
	\begin{center}
	\begin{tabular}{cc}
		\includegraphics[width=.4\textwidth]{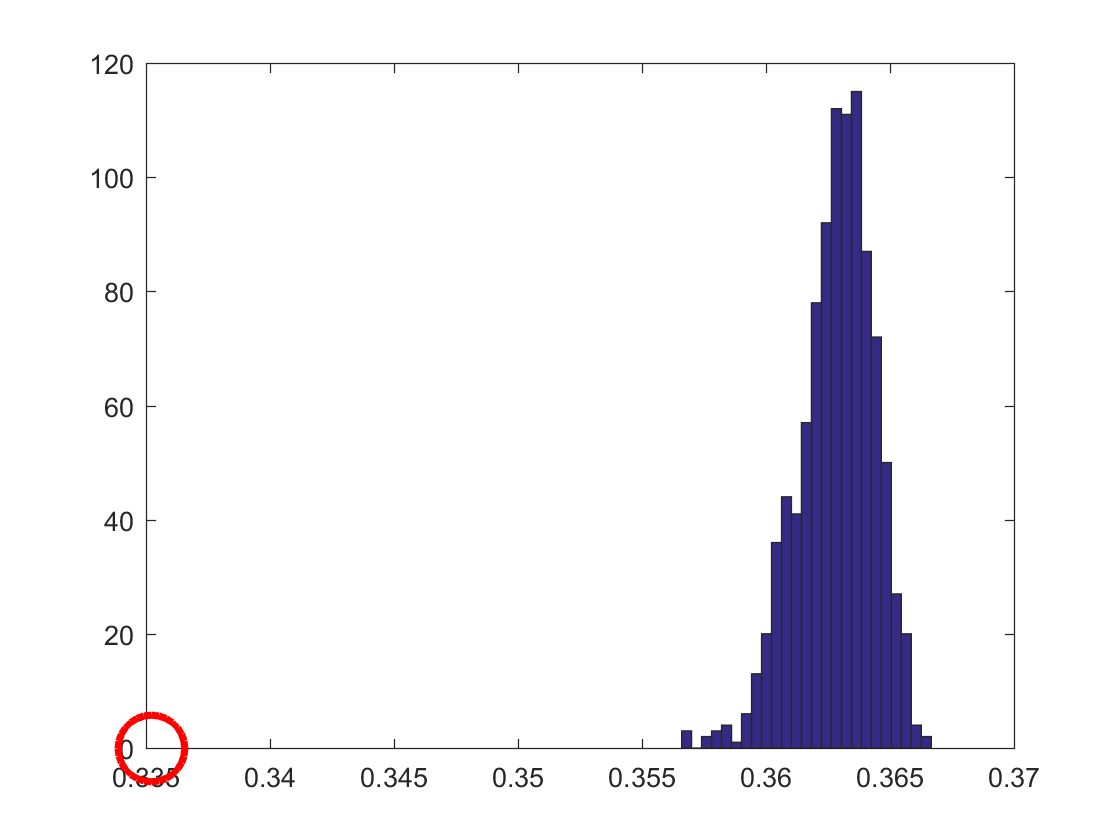} & 
		\includegraphics[width=.4\textwidth]{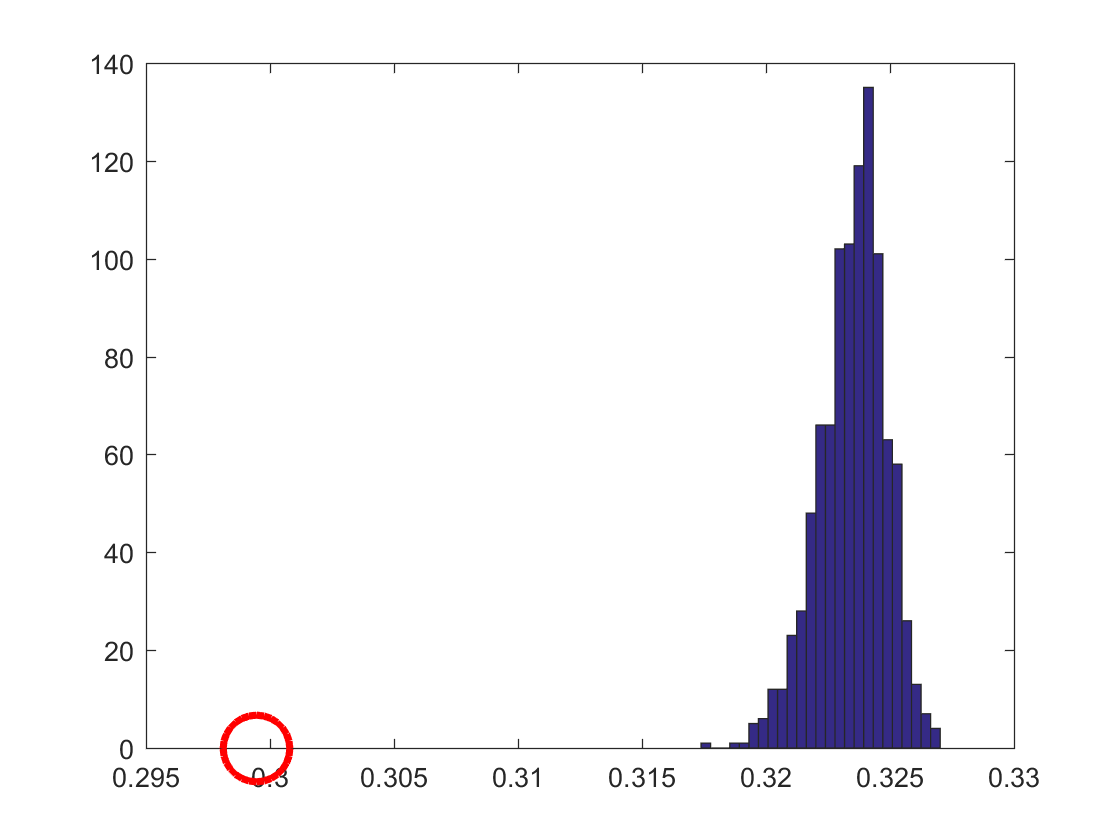}\\
		Percent in SNAP & Percent with Bachelors degree or higher
	\end{tabular}
\end{center}
\caption{The red circle represents the true value of $error(f)$, whereas the blue bars are the histogram of values of $error(\widetilde{f})$ for 1000 random permutations. $K_{embedding}$ computed with $\sigma=0.5$.}\label{fig:permutation}
\end{figure}

Figure \ref{fig:permutation} shows the comparison of $error(f_{SNAP})$ to 1000 of its permutations, and similarly for $error(f_{Bachelors})$.  In both cases, the true value of $error(f)$ is significantly smaller than any of 1000 random permutations of the values.  In other words, the correlation between the county embedding and $f_{SNAP}$ is strong enough that, if we were to randomly shuffle all the counties' placements in the embedding 1000 times, none of those organizations would exhibit the level of prediction of $f_{SNAP}$ that our organization produces.

%\begin{figure}[!h]
%	\footnotesize
%	\begin{center}
%		\begin{tabular}{ccc}
%			\includegraphics[width=.25\textwidth]{embeddingWBI_small.png} & 
%			\includegraphics[width=.25\textwidth]{embeddingCommunity_small.png} & 
%			\includegraphics[width=.25\textwidth]{embeddingSocial_small.png} \\
%			WBI Score & Community Score & Social Score \\
%			\includegraphics[width=.25\textwidth]{embeddingPurpose_small.png} & 
%			\includegraphics[width=.25\textwidth]{embeddingPhysical_small.png} & 
%			\includegraphics[width=.25\textwidth]{embeddingFinancial_small.png} \\
%			Purpose Score & Physical Score & Financial Score 
%		\end{tabular}
%	\end{center}
%	\caption{}\label{fig:factorColors_small}
%\end{figure}

In addition, we can use the residual $f - K_{embedding}f$ to identify outlier counties.  This would be a county $Z_i$ that has, for example, a similar well-being profile to counties with high educational achievement, though the rate of educational achievement in $Z_i$ differs significantly.  These types of counties can be easily identified and become the subject of further study.  This means we can now identify counties that have populations with similar experiences of well-being, but varying demographic characteristics.  This outlier information could also be applied to county-level health outcomes and other types of county level information.

\section{Conclusion}
We used bigeometric organization to create a distance metric between any two data points $x_i, x_j$, and used that metric to construct an approximate earth mover's distance to estimate distances between any pair of distribution classes $Z_i, Z_j$.  We use the eigenfunctions of the kernel on the network of pairs of classes to define a low-dimensional embedding of the distances between classes.  Finally, we use this construction to create a low-dimensional map of all counties studied in the Gallup-Healthways Well-Being Index survey, where the map is constructed from people's similarity of responses to the survey.  

This map gives a novel coordinate space by which to consider counties, as well as a simple visualization of their similarity.  This survey map can be broken into regions and begin to separate out, for example, different types of counties that have a similar overall GHWBI score.
This also allows one to easily identify outlying counties that lie in a region of the survey map that does not reflect other descriptions of the county, such as census data or aggregated medical histories.  These counties are of particular interest and can be marked for further examination.

\section*{Acknowledgments}
The authors would like to thank Jack Welsh, Elizabeth Rula, Kennith Kell, and Raphy Coifman, as well as Gallup and Healthways for access to the data.  Alexander Cloninger is partially supported by NSF grant DMS-1402254.

%\newpage
\bibliographystyle{plain}

\end{document}